\documentclass[lettersize,journal]{IEEEtran}
\usepackage{amsmath,amsfonts}
\usepackage{algorithm}
\usepackage{algorithmic}
\usepackage{array}
\usepackage{textcomp}
\usepackage{tcolorbox}
\usepackage{bbding}
\usepackage{stfloats}
\usepackage{url}
\usepackage{verbatim}
\usepackage{graphicx}
\usepackage{cite}
\usepackage{multirow}
\usepackage{booktabs}
\usepackage{makecell}
\usepackage{array}
\usepackage{graphicx}
\usepackage{tabularx}
\usepackage{subcaption}
\usepackage{caption}
\usepackage{color}

\begin{document}

\title{Semantic Communication based on Large Language Model for Underwater Image Transmission}

\author{Weilong Chen, Wenxuan Xu, Haoran Chen, Xinran Zhang, Zhijin Qin,~\IEEEmembership{Senior Member,~IEEE},  Yanru Zhang,~\IEEEmembership{Member,~IEEE}, and Zhu Han,~\IEEEmembership{Fellow,~IEEE}


\thanks{This work has been submitted to the lEEE for possible publication.Copyright may be transferred without notice, after which this version may no longer be accessible.}
}


\maketitle

\begin{abstract}
Underwater communication is essential for environmental monitoring, marine biology research, and underwater exploration. Traditional underwater communication faces limitations like low bandwidth, high latency, and susceptibility to noise, while semantic communication (SC) offers a promising solution by focusing on the exchange of semantics rather than symbols or bits. However, SC encounters challenges in underwater environments, including semantic information mismatch and difficulties in accurately identifying and transmitting critical information that aligns with the diverse requirements of underwater applications. To address these challenges, we propose a novel Semantic Communication (SC) framework based on Large Language Models (LLMs). Our framework leverages visual LLMs to perform semantic compression and prioritization of underwater image data according to the query from users. By identifying and encoding key semantic elements within the images, the system selectively transmits high-priority information while applying higher compression rates to less critical regions. On the receiver side, an LLM-based recovery mechanism, along with Global Vision ControlNet and Key Region ControlNet networks, aids in reconstructing the images, thereby enhancing communication efficiency and robustness. Our framework reduces the overall data size to 0.8\% of the original. Experimental results demonstrate that our method significantly outperforms existing approaches, ensuring high-quality, semantically accurate image reconstruction.
\end{abstract}


\section{Introduction}
\subsection{Motivation and Background}
Underwater communication is a critical technology with widespread applications in various domains, such as environmental monitoring, marine biology research, and underwater exploration \cite{ali2020recent}. The ability to transmit multimodal data efficiently and reliably in underwater environments is essential for collecting and sharing vital information. For instance, environmental monitoring relies on underwater sensors and vehicles to track changes in marine ecosystems, detect pollution, and study climate change impacts. Marine biologists use underwater communication systems to observe and understand marine life behaviors, migrations, and interactions \cite{jaafar2022overview}. Additionally, underwater exploration missions, including resource extraction and archaeological studies, depend heavily on effective communication systems to ensure the safety and success of operations. The integration and transmission of multimodal data, such as images, videos, and sensory data, are crucial to providing comprehensive insights.

Traditional underwater communication techniques primarily include acoustic communications, optical communications, and radio frequency (RF) communications~\cite{zhu2020recent}. Acoustic communication is the most widely used and practical underwater communication method, capable of providing wide coverage up up to tens of kilometers~\cite{zhu2023internet}. However, the acoustic communication system is facing several issues. First, limited bandwidth results in low data transmission rates on the order of kbps, limiting the amount of information that can be sent in a given time. Secondly, high latency is another issue, as the speed of sound in water is much slower compared to electromagnetic waves in air~\cite{gauni2021design}, thus leading to delays in data transmission. Moreover, underwater acoustic signals are highly susceptible to noise from marine life, human activities, environmental factors such as temperature gradients and salinity variations\cite{tang2020research}, signal attenuation and multipath effects\cite{menaka2022challenges}, which could degrade signal quality and lead to data distortion. Therefore, in the complex underwater physical environment, traditional acoustic communication technologies face challenges such as low bandwidth, high latency, significant attenuation, and low robustness. 


Currently, some researchers utilize Semantic Communication (SC) based on artificial intelligence to address communication problems under low bandwidth and significant attenuation\cite{farsad2018deep,xie2021deep,xu2023deep,jiang2024large,qin2021semantic}. By introducing a semantic channel to the communication system\cite{iyer2023survey}, SC extracts and encodes the semantic information of data at the transmitter end, transmits it through the semantic channel, and successfully decodes the semantic information at the receiver end\cite{luo2022semantic}. Traditional communication systems focus solely on the accuracy of bitstream transmission, ignoring the implicit semantic information behind the data. In contrast, SC can extract semantic information, achieving accurate semantic transmission\cite{shi2021semantic} and reducing the amount of transmitted data, thereby optimizing the use of limited bandwidth.


However, traditional SC faces significant challenges in underwater environments, particularly for semantic information tasks. {Firstly, the delay spread from multipath propagation and Doppler shifts from relative motion leads to \textit{semantic information mismatch}.} Multipath effects and frequency-selective fading distort signals, making it difficult for traditional SC to accurately extract and transmit semantic content. The Doppler effect further exacerbates this by introducing frequency shifts that disrupt the coherence of semantic data. {Traditional methods, which rely heavily on a singular recovery mechanism without semantic coherence enhancement, are insufficient to accurately reconstruct meaningful data amidst the complex noise and severe interference of underwater environments.}

Secondly, SC faces the challenge of \textit{accurately identifying and transmitting different critical information in alignment with the diverse requirements of underwater applications}. {The traditional SC framework typically lacks the flexibility needed to distinguish and prioritize varying types of data within communication channels. As a result, they struggle to effectively manage the complexity of underwater applications, where different semantic content must be accurately identified, prioritized, and transmitted according to the specific requirements of the task at hand.} For instance, environmental monitoring may focus on transmitting images of coral reef health, while marine biology might prioritize detecting specific marine species. Non-essential semantic information can interfere with the transmission of important data, potentially leading to misinterpretations or delays that negatively impact underwater operations and decision-making processes.


\subsection{Related Work}
 
\subsubsection{Underwater Communication}
Underwater communication technique plays an important role in marine biology observation, oil and gas drilling exploration, and natural disaster monitoring. Traditional underwater wireless communication technologies primarily include acoustic communication, optical communication, and radio frequency communication. Acoustic communication utilizes sound signals for transmission, allowing for the greatest propagation distances, which is currently the dominant technology in underwater communication. In the unique physical environment underwater, acoustic signals are significantly affected by multipath interference and ambient noise. Equalization techniques are employed to overcome inter symbol interference  (ISI) caused by multipath propagation\cite{pranitha2020analysis,feng2023ofdm}. Acoustic communication is characterized by limited bandwidth and slow data transmission rates. Modems are employed to improve transmission rates, mitigate multipath interference, and enhance noise resistance\cite{huang2020adaptive,zia2021state}. To reduce data redundancy, data compression techniques are widely used in acoustic communication, including compressed sampling\cite{wu2018compressive} and compressed sensing technologies\cite{tabata2020improvement}. Moreover, acoustic signal propagation is easily affected by underwater physical conditions such as temperature, pressure, salinity, and water density, which could lead to signal attenuation and data distortion. Optical and RF communication can provide higher data rates compared to acoustic communication, but both methods are limited in terms of transmission distance\cite{sun2020review}. Optical signals exhibit exponential attenuation in water, moreover, optical communication requires a line-of-sight connection between the transmitter and receiver\cite{ali2022recent}, which can result in high error rates and data loss in water with a high concentration of particulates. RF technology can provide medium data transmission rates, but electromagnetic signals experience significant attenuation in water\cite{aboderin2017performance}, limiting the long-distance propagation of RF technology. Therefore, existing traditional underwater communication technologies face several significant challenges, including severe attenuation, multipath effects, dispersion, limited bandwidth, low transmission rates, and low robustness.

\subsubsection{Semantic Communication}
Thanks to the development of artificial intelligence and communication technologies, researchers have developed SC systems based on deep learning to help address the limited bandwidth and robustness of communication.
SC can extract the meaning of data in the semantic domain, filter out irrelevant and unimportant information, and significantly reduce the data volume by further compressing the semantic information while preserving the meaning. Therefore, when bandwidth is limited or the signal-to-noise ratio (SNR) is relatively low, SC systems may still perform well, maintaining robustness and reliability.
Xie \textit{et al.} proposed a Transformer-based SC framework to maximize system capacity and eliminate semantic errors\cite{xie2021deep} while Wang \textit{et al.} propose the LLM-SC framework that integrates Large Language Models directly into the physical layer of semantic communication systems for text data only\cite{wang2024large}.
However, compared to text, image data is semantically richer and more bandwidth-sensitive, making the transmission of image data more challenging.
Huang \textit{et al.} proposed an image semantic encoding method based on Generative Adversarial Networks (GANs)\cite{huang2021deep}.
Wu \textit{et al.} proposed a semantic transmission system for segmenting regions of interest in images\cite{wu2023semantic}.
Eirina \textit{et al.} proposed an image transmission method based on joint source-channel coding\cite{bourtsoulatze2019deep}.
Cao \textit{et al.} proposed an underwater Metaverse framework utilizing quantum imaging and generative AI to create highly immersive and realistic virtual underwater environments~\cite{cao2024unified} while Jiang \textit{et al.} propose a LAM-based SC framework for image data that uses AI models to segment and encode semantic data automatically~\cite{jiang2024large}. Zhang \textit{et al.} also propose a generative AI-aided SC framework that integrates deep generative models for semantic feature extraction and reconstruction~\cite{zhang2024semantic}.
However, these methods cannot identify the different critical information that aligns with the diverse requirements of underwater applications, which can lead to the loss of critical information and reduced transmission efficiency. Additionally, multipath propagation and Doppler spread from relative motion in underwater environments further intensify the difficulties in maintaining robust data transmission.

Therefore, it is imperative to adequately consider identifying the varying important information of different tasks when designing SC systems.  Moreover, we need to further consider the semantic information mismatch caused by the complex underwater environments, ensuring that critical information is preserved and accurately reconstructed even under challenging underwater conditions.

\subsection{Our Contributions}
To address these challenges, we propose a novel SC framework based on Large Language Models (LLMs). {Unlike traditional models, LLMs excel in their generalization capability, allowing them to be readily applicable across a broad range of tasks within only one model. Beyond this, they can also interpret user intentions to distinguish and prioritize different types of information based on their importance. This enables LLMs to selectively filter and manage critical data according to the specific needs of the task at hand, offering a significant advantage over traditional models.} Our framework leverages visual LLMs~\cite{wang2024visionllm} to comprehend queries from individuals above the surface and perform semantic compression and prioritization of image data, aligning with the diverse requirements of underwater applications. This method allows for the identification and encoding of key semantic elements within images and selectively transmits high-priority information while applying higher compression rates to less critical regions, enhancing the effectiveness and relevance of data transmission. On the receiver side, we employ a text LLM recovery mechanism to reconstruct the textual data, and two new ControlNet networks to assist the diffusion model in recovering the image to the original size. {Those recovery mechanisms enhance the semantic coherence to mitigate the information mismatch}. This approach not only reduces the overall data size to 0.8\% of the original, but also significantly improves the resilience of the communication system against noise and signal loss.

The main contributions of this paper are summarized as follows:

\begin{itemize}
    \item We introduce an innovative LLM-based SC framework to enhance underwater communication. This framework can perform semantic compression and prioritization, enhancing data transmission by adapting dynamically to the different requirements of underwater applications.
    
    \item {The proposed framework integrates two newly designed ControlNet networks and a text-based LLM recovery mechanism with a diffusion model, enhancing the semantic coherence while effectively mitigating information mismatch.}
    
    \item Our experimental results show that our framework can compress data to just 0.8\% of its original size and still maintain high-quality transmission effects even under significant noise conditions.
    
\end{itemize}

The remainder of this paper is organized as follows. Section 2 provides a review of related work in underwater communication and semantic compression techniques. Section 3 describes the proposed framework in detail, including its architecture and technical components. Section 4 presents the experimental setup and results, highlighting the performance improvements achieved by our method. Finally, Section 5 concludes the paper and discusses potential directions for future research.

\section{System Model}\label{section3}

\begin{figure*}[ht]
    \centering
  \includegraphics[width=0.9\textwidth]{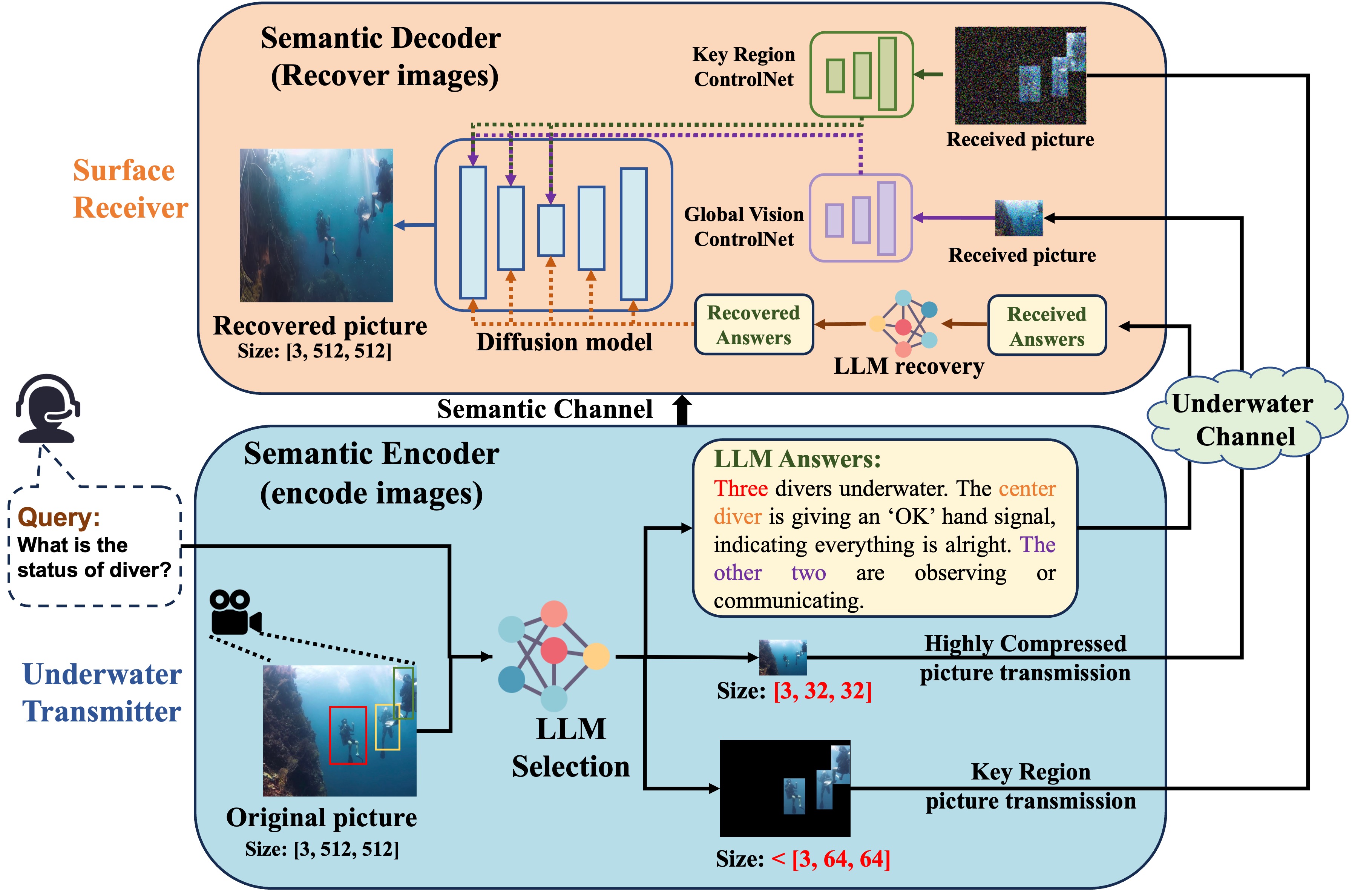}
  
  \caption{The framework structure of proposed LLM-based semantic communication.}
  \label{fig:framework_structure}
  \vspace{-2 mm}
\end{figure*}

The proposed comprehensive framework is tailored specifically for underwater image transmission, which is shown in Fig. \ref{fig:framework_structure}. This framework initiates with a query generated by individuals located above the water surface, which is subsequently transmitted to the underwater environment. Upon receiving the query, the underwater transmitter employs a semantic encoder integrated with a LLM-based information prioritization mechanism. This mechanism is designed to identify and prioritize critical parts of the visual information based on the query's context. The essential information is then compressed into a format suitable for transmission through the underwater communication channel. Once transmitted, the information is received and decoded by a semantic decoder with a diffusion model and LLM recovery at the receiver's end. The decoded visual information is subsequently reconstructed and presented to the individuals above the water. This approach ensures efficient and effective transmission of visual information in underwater communication scenarios.

\subsection{Semantic Encoder}
Initially, an individual on the surface provides a query specifying the desired visual information. This query is then transmitted to the semantic encoder by the wireless channel. The semantic encoder is composed of two primary components: LLM-based Information Prioritization and Image Highly Compression.

\subsubsection{LLM-based Information Prioritization} To efficiently transmit images underwater and extract their semantic information, we employ visual LLM (vLLM) for information prioritization. The process begins with the reception of a query from individuals above the water surface. These queries, typically articulated in natural language, must be accurately understood and interpreted to extract the relevant contextual information.

Firstly, given the query $\boldsymbol{q}$ and image $\boldsymbol{i} \in \mathbb{R}^{C \times H \times W}$ taken underwater, the vLLM $\mathcal{F}_l$ generates an answer and identifies the key regions of the original image. This process can be described as:
\begin{equation}
    \boldsymbol{a} = \mathcal{F}_{l}(\boldsymbol{q}, \boldsymbol{i}), 
\end{equation}
where \(\boldsymbol{a}\) represents the answer provided by the vLLM. Subsequently, \(\boldsymbol{a}\) is inputted into \(\mathcal{F}_l\) along with \(\boldsymbol{i}\) to determine the key informational regions of the image, as follows:

\begin{equation}
    \text{bbox}_{k} = \mathcal{F}_{l}(\boldsymbol{a}, \boldsymbol{i}), 
\end{equation}
where \(\text{bbox}_{k}^{n}\) denotes the bounding box encompassing the key regions. This bounding box is then applied to the original image \(\boldsymbol{i}\), followed by image compression techniques to resample and compress the image, yielding \(\boldsymbol{i}_{k} \in \mathbb{R}^{C^{k} \times H^{k} \times W^{k}}\). In comparison to \(\boldsymbol{i}\), \(\boldsymbol{i}_{k}\) contains only the essential information. Specifically, \(C^{k} < C\), \(H^{k} < H\), and \(W^{k} < W\), which helps reduce the image size. Additionally, since only the key regions are transmitted, the data size can be further minimized within the original \(\mathbb{R}^{C^{k} \times H^{k} \times W^{k}}\) dimensions, ensuring an even lower transmission overhead. This selective extraction results in a representation that, while retaining the full dimensionality of the original image, significantly reduces informational entropy by excluding extraneous data and preserving only the semantically relevant content.

\subsubsection{Image Highly Compression}To efficiently transmit the entire image with minimal communication bandwidth, we employ image compression techniques to resample and compress the complete image. Resampling involves altering the image resolution, effectively reducing the number of pixels and, consequently, the overall data size. By decreasing the spatial dimensions, we minimize the amount of information that needs to be transmitted, thus reducing bandwidth requirements. Compared to compressing only the non-key regions, this holistic approach offers superior benefits because the image is uniformly compressed. Therefore, it is more efficient to apply compression directly to the entire image rather than separately compressing non-key regions. This strategy ensures that the overall data size is minimized while maintaining an acceptable level of visual quality throughout the image. The highly compressed image \(\boldsymbol{i}_{hc}\) can be expressed as:
\begin{equation}
    \boldsymbol{i}_{hc} = \text{RESAMPLE}(\boldsymbol{i}), 
\end{equation}
where \(\boldsymbol{i}_{hc} \in \mathbb{R}^{C^{hc} \times H^{hc} \times W^{hc}}\). Here, \(C^{hc}, H^{hc},\) and \(W^{hc}\) are significantly smaller than \(C^{ci}, H^{ci},\) and \(W^{ci}\), resulting in a substantial reduction in size during transmission.

The elements \(\boldsymbol{a}\), \(\boldsymbol{i}_{k}\), and \(\boldsymbol{i}_{hc}\) are then transmitted from the semantic encoder to the semantic decoder through the physical channel.

\subsection{Underwater Acoustic Communication Channel} 
In this section, a realistic physical communication environment is considered. In a shallow water acoustic propagation environment, to consider the multipath propagation and Doppler spread, we first define the path loss between surface and underwater robot, which can be defined as:
\begin{equation}
A\left(d, f\right)=d^{\sigma} a(f)^{d},
\end{equation}
where $\sigma$ is the spread factor, $d$ is the distance  and $a(f)$ is the absorption coefficient, as calculated using the Thorp formula \cite{9978925},  measured by dB/km with $f$ in kHz, which is expressed as:
\begin{equation}
\begin{aligned}
10 \log (a(f))= & 0.11 \frac{f^2}{1+f^2}+44 \frac{f^2}{4100+f^2} \\
& +2.75 \times 10^{-4} f^2+0.003 .
\end{aligned}
\end{equation}

Secondly, according to \cite{7006722}, we consider the underwater environmental noise $\mathcal{N}(f)$ which is composed of turbulence, ship, wind and thermal noises \cite{9068243}, denoted as $\mathcal{N}_1(f),\mathcal{N}_2(f),\mathcal{N}_3(f),\mathcal{N}_4(f)$, respectively. This can be defined as:
\begin{equation}
\mathcal{N}(f)=\mathcal{N}_1(f)+\mathcal{N}_2(f)+\mathcal{N}_3(f)+\mathcal{N}_4(f),
\end{equation}
and those components can be described as:
\begin{equation}
\left\{\begin{array}{l}
10 \log \mathcal{N}_1(f)=17-30 \log f, \\
10 \log \mathcal{N}_2(f)=30+20 s+\log \left(f^{26} /(f+0.03)^{60}\right), \\
10 \log \mathcal{N}_3(f)=50+7.5 v^{1 / 2}+20 \log \left(f /(f+0.4)^2\right), \\
10 \log \mathcal{N}_{4}(f)=-15+20 \log f,
\end{array}\right.
\end{equation}
where $s$ is the shipping activity factor between $[0,1]$, and $v$ is the wind speed measured in m/s. The channel conditions are characterized by the signal-to-noise ratio (SNR), defined as:
\begin{equation}
\text{SNR} = \frac{P_\text{signal}}{P_\text{noise}},
\end{equation}
where \(P_{\text{signal}}\) and \(P_{\text{noise}}\) represent the average received signal and noise powers, respectively. The $P_{\text{noise}}$ is decided by the $\mathcal{N}(f) \cdot A\left(d, f\right)$. Notably, the aggregate size of \(\boldsymbol{i}_{k}^{\prime}\), \(\boldsymbol{a}^{\prime}\), and \(\boldsymbol{i}_{hc}^{\prime}\) constitutes only 0.8\% of the original image \(\boldsymbol{i}\) in our experiments, underscoring the bandwidth efficiency of our proposed method.


\subsection{Semantic Decoder}
Upon receiving \(\boldsymbol{i}_{k}^{\prime}\), \(\boldsymbol{i}_{hc}^{\prime}\), and \(\boldsymbol{a}^{\prime}\), the receiver employs a diffusion model \(\mathcal{F}_{d}\) to reconstruct the original image. The semantic decoder in the receiver leverages two ControlNets \cite{zhang2023adding} and is guided by textual information to ensure accurate and detailed image recovery.

The diffusion model \(\mathcal{F}_{d}\) belongs to a class of generative models that define a forward and reverse process to generate data. The forward process incrementally adds noise to the data, while the reverse process denoises the data to reconstruct the original input. Formally, the forward process is represented as:
\begin{equation}
q(\mathbf{x}_{1:T} | \mathbf{x}_0) = \prod_{t=1}^T q(\mathbf{x}_t | \mathbf{x}_{t-1}),
\end{equation}
where \(\mathbf{x}_0\) is the original data and \(\mathbf{x}_t\) represents the noisy data at time step \(t\). The reverse process aims to recover \(\mathbf{x}_0\) from \(\mathbf{x}_T\) by progressively denoising the data:
\begin{equation}
p_\theta(\mathbf{x}_{0:T}) = p(\mathbf{x}_T) \prod_{t=1}^T p_\theta(\mathbf{x}_{t-1} | \mathbf{x}_t),
\end{equation}
where \(p_\theta\) represents the learned reverse process parameterized by \(\theta\). 

To better recover the key regions, we proposed the Key Region ControlNet $\mathcal{C}_{k}$, which is dedicated to processing \(\boldsymbol{i}_{k}^{\prime}\). This network is specifically designed to enhance and accurately generate the critical regions identified during the LLM-based Information Prioritization phase, which can be referred to as:
\begin{equation}
\boldsymbol{f}_k = \mathcal{C}_{k}(\boldsymbol{i}_{k}^{\prime}),
\end{equation}
where $\boldsymbol{f}_k$ is the output of the $\mathcal{C}_{k}$ and will be integrated into the denoising process in the diffusion model. By focusing on these key areas, the Key Region ControlNet ensures that the significant details, such as semantic segmentation and human pose estimation, are preserved and reconstructed with high fidelity. This targeted enhancement allows for a more refined generation of the most important parts of the image, thereby maintaining the integrity of the prioritized information.

To recover the rest restoration of the image without emphasizing fine details, we proposed Global Vision ControlNet, which is responsible for handling \(\boldsymbol{i}_{ci}^{\prime}\).  The Global Vision ControlNet aims to reconstruct the broader spatial layout and general color distribution of the image, which can be described as:
\begin{equation}
\boldsymbol{f}_{ci} = \mathcal{C}_{ci}(\boldsymbol{i}_{ci}^{\prime}),
\end{equation}
where $\boldsymbol{f}_{ci}$ is the output of the $\mathcal{C}_{ci}$ and will be also integrated into the denoising process in the diffusion model. The Global Vision ControlNet ensures that the reconstructed image retains the correct positioning and overall appearance of the objects within it. This broader approach helps in maintaining the coherence of the entire image while ensuring that the critical regions, processed by the Key Region ControlNet, are accurately integrated into the final output.

In addition to the visual information, we also employ an LLM-based recovery process for the textual guide \(\boldsymbol{a}^{\prime}\). We also utilize the $\mathcal{F}_l$ as the base model of the LLM recovery process. To recover the text, we use the prompt:

\begin{tcolorbox}
Help me recover the sentence from the noise sentence. The sentence is \(\boldsymbol{a}^{\prime}\).
\end{tcolorbox}
This process can be defined as:
\begin{equation}
\bar{\boldsymbol{a}} = \mathcal{F}_l(\text{prompt}, \boldsymbol{a}^{\prime}),
\end{equation}
where \(\bar{\boldsymbol{a}}\) is the final text utilized by the diffusion model. This step is crucial for preserving the semantic context and ensuring that the textual information is accurately reconstructed. The LLM-based textual recovery leverages the advanced capabilities of large language models to interpret and restore the original textual content, thereby enhancing the overall quality and relevance of the reconstructed information.

To provide fine-grained control over the generative process of diffusion models, we incorporate three guidance signals to enable the generation of more precise and contextually relevant outputs by conditioning the diffusion model on additional information. These guidance signals can be integrated into the reverse diffusion process as follows:
\begin{equation}
p_\theta(\mathbf{x}_{t-1} | \mathbf{x}_t, \mathbf{c}) = \mathcal{N}(\mathbf{x}_{t-1}; \mu_\theta(\mathbf{x}_t, \mathbf{c}, t), \Sigma_\theta(\mathbf{x}_t, \mathbf{c}, t))
\end{equation}
where \(\mathbf{c}\) represents the control signal provided by the three guidance components, and \(\mu_\theta\) and \(\Sigma_\theta\) are the mean and covariance functions parameterized by \(\theta\). \(\mathbf{c}\) is defined as:

\begin{equation}
\mathbf{c} = \left\{  \bar{\boldsymbol{a}},\left[\alpha \boldsymbol{f}_k + (1-\alpha)\boldsymbol{f}_{ci}\right] \right\},
\end{equation}
where the hyperparameter \(\alpha\) controls the balance between the contributions of the Key Region ControlNet and the Global Vision ControlNet. The complete process of the diffusion model \(\mathcal{F}_{\theta}\), which integrates the outputs from the two ControlNets and the LLM-based textual recovery to generate the final image, can be defined as:
\begin{equation}
\boldsymbol{i}_{r} = \mathcal{F}_{\theta}\left(\mathbf{x}_t,\mathbf{c} \right),
\end{equation}
where \(\boldsymbol{i}_{r} \in \mathbb{R}^{C \times H \times W}\) is the recovered image. By combining these three guidance signals, the diffusion model effectively synthesizes the visual content, ensuring that the reconstructed image is both semantically accurate and visually coherent. The integration of these components allows the model to leverage the strengths of each guidance signal, resulting in a comprehensive and high-quality reconstruction of the original image. The whole process can be found in Alg. \ref{alg:LLM_SC}.

\subsection{Training Process}
To train the semantic encoder, there are question-answering loss and key information prioritization loss. The question answering loss is designed to ensure that the vLLM's answer \(\boldsymbol{a}\) accurately reflects the text label \(\boldsymbol{y}\). This is achieved by using a cross-entropy loss between the predicted answer and the true label:

\begin{equation}
    \mathcal{L}_{a} = -\frac{1}{N} \sum_{n=1}^{N} \boldsymbol{y}^{(n)} \log \boldsymbol{a}^{(n)},
\end{equation}
where $N$ is the size of the training data. This loss ensures that the semantic information provided by the LLM is accurately captured and maintained during the reconstruction process.

The key information prioritization loss is designed to ensure that the extracted key regions match the original key regions as closely as possible. To achieve this, we use the bounding box-based loss to ensure that the spatial alignment and dimensions of the key regions are preserved. The bounding box loss is defined as the sum of the Intersection over Union (IoU) loss for the bounding boxes in the original and reconstructed images:

\begin{equation}
    \mathcal{L}_{key, bbox} = \frac{1}{N} \sum_{n=1}^{N} \left(1 - \text{IoU}(\text{bbox}_{k}^{n}, \hat{\text{bbox}_{k}^{n}})\right),
\end{equation}
where \(\hat{\text{bbox}_{k}^{n}}\) and \(\text{bbox}_{k}^{n}\) are the bounding boxes of the key regions in the original and extracted images, respectively. 

To train the semantic decoder, two reconstruction controlnet losses are utilized. For the Key Region ControlNet, Given a set of conditions including time step $t$, answer $\bar{\boldsymbol{a}}$, as well as a task-specific condition $\boldsymbol{f}_k$, Key Region ControlNet learn a network $\theta_k$ to predict the noise added to the noisy image $\mathbf{x}_t$ with:
\begin{equation}
\left.\mathcal{L}_k=\mathbb{E}_{\boldsymbol{i}, t, \bar{\boldsymbol{a}}, \boldsymbol{f}_k, \epsilon \sim \mathcal{N}(0,1)}\left[\| \epsilon-\epsilon_{\theta_k}\left(\mathbf{x}_t, t, \bar{\boldsymbol{a}}, \boldsymbol{f}_k\right)\right) \|_2^2\right],
\end{equation}
where $\mathcal{L}_k$ is the overall learning objective of the diffusion model. This learning objective is directly used in fine-tuning diffusion models with Key Region ControlNet. The loss of Global Vision ControlNet can be also written as:
\begin{equation}
\left.\mathcal{L}_{ci}=\mathbb{E}_{\boldsymbol{i}, t, \bar{\boldsymbol{a}}, \boldsymbol{f}_{ci}, \epsilon \sim \mathcal{N}(0,1)}\left[\| \epsilon-\epsilon_{\theta_{ci}}\left(\mathbf{x}_t, t, \bar{\boldsymbol{a}}, \boldsymbol{f}_{ci}\right)\right) \|_2^2\right],
\end{equation}
where $\theta_{ci}$ is the network parameters of Global Vision ControlNet.

\begin{figure}[!t]
    \begin{algorithm}[H]
        \caption{Proposed LLM-based Semantic Communication}
        \label{alg:LLM_SC}
        \begin{algorithmic}[1]
            \REQUIRE A query $\boldsymbol{q}$, underwater image $\boldsymbol{i}$, pretrained vLLM model $\mathcal{F}_{l}$, Key Region ControlNet network $\mathcal{C}_{k}$ with initialized parameters $\theta_{k}$, Global Vision ControlNet network $\mathcal{C}_{hc}$ with initialized parameters $\theta_{hc}$, diffusion model $\mathcal{F}_{\theta}$ with pretrained parameters $\theta$.
        \end{algorithmic}
        
        \textbf{Transmitter Operations:}
        \begin{algorithmic}[1]
            \STATE Compute \(\boldsymbol{a}\), \(\text{bbox}_{k}\), and \(\boldsymbol{i}_{k}\) using \(\mathcal{F}_{l}\) according to (1) and (2).
            \STATE Resample and compress \(\boldsymbol{i}\) to obtain \(\boldsymbol{i}_{hc}\) using (3).
            \STATE Transmit \(\boldsymbol{a}\), \(\boldsymbol{i}_{k}\), and \(\boldsymbol{i}_{hc}\) through the physical channels to the receiver as per (4), (5), and (6).
        \end{algorithmic}

        \textbf{Receiver Operations:}
        \begin{algorithmic}[1]
            \STATE Receive \(\boldsymbol{a}^\prime\), \(\boldsymbol{i}_{k}^\prime\), and \(\boldsymbol{i}_{hc}^\prime\).
            \STATE Compute \(\boldsymbol{f}_k\) using \(\mathcal{C}_{k}\) according to (10).
            \STATE Compute \(\boldsymbol{f}_{hc}\) using \(\mathcal{C}_{hc}\) according to (11).
            \STATE Recover the text from \(\boldsymbol{a}^\prime\) using \(\mathcal{F}_{l}\) according to (12) to obtain \(\bar{\boldsymbol{a}}\).
            \STATE Perform the diffusion process and recover the entire image using the diffusion model \(\mathcal{F}_{\theta}\) with \(\boldsymbol{f}_k\), \(\boldsymbol{f}_{hc}\), and \(\bar{\boldsymbol{a}}\).
            \STATE Obtain the recovered image \(\boldsymbol{i}_{r}\).
        \end{algorithmic}   
        
    \end{algorithm}
\end{figure}

\section{Case Study}\label{section4}

Here, we present a comprehensive analysis of the performance of the proposed method on the real-world underwater dataset named SUIM Dataset \cite{islam2020semantic}. Specifically, we aim to perform 3 evaluations: (\textbf{E1}) How does the proposed method outperform other approaches? (\textbf{E2}) How does each part contribute to the performance? (\textbf{E3}) How do critical hyperparameters and components impact the proposed method?

\subsection{Dataset Description and Evaluation Metrics}

To evaluate the performance of our proposed method, we utilize the SUIM Dataset \cite{islam2020semantic}, which comprises over 1,500 images annotated for eight object categories: fish (vertebrates), reefs (invertebrates), aquatic plants, wrecks/ruins, human divers, robots, and seafloor. Based on these categories, we formulate various questions pertaining to the object categories within the images. For the key region identification, we convert the segmentation annotations to bounding boxes corresponding to the object categories. The whole dataset is divided into 80\% for training and 20\% for testing.

To rigorously evaluate the performance of our proposed method, we employ several metrics: Fréchet Inception Distance (FID) \cite{heusel2017gans}, Structural Similarity Index (SSIM) \cite{wang2004image}, Contrastive Language-Image Pretraining (CLIP) \cite{radford2021learning}, and Learned Perceptual Image Patch Similarity (LPIPS) \cite{zhang2018perceptual}. FID measures the similarity between the generated images and real images by comparing the distributions of their feature representations extracted from a pretrained Inception network. FID can be defined as:
\begin{equation}
    d_F^2(X,Y) = \Vert \mu_X - \mu_Y \Vert _2^2 + Tr\left[\Sigma_X + \Sigma_Y - 2(\Sigma_X\Sigma_Y)^{\frac{1}{2}}\right],
\end{equation}
where $X$ and $Y$ are the distributions of real image and generated image, respectively, $\mu_X$ and $\mu_Y$ are the means, $\Sigma_X$ and $\Sigma_Y$ are the covariance of $X$ and $Y$, a lower FID score indicates higher quality and more realistic images. 

SSIM assesses the structural similarity between the original and reconstructed images. It evaluates the perceived quality based on luminance, contrast, and structure. SSIM can be defined as:
\begin{equation}
    \text{SSIM}(x,y)= [l(x,y)]^\alpha[c(x,y)]^\beta[s(x,y)]^\gamma,
\end{equation}
where $l$ is the luminance, $c$ is the contrast and $s$ is the structure comparison between the original image $x$ and reconstructed image $y$. The exponent $\alpha, \beta, \gamma$ are positive constants. SSIM values range from -1 to 1, with higher values indicating better similarity.

The luminance, contrast, and structure comparison of two images can be expressed respectively as:
\begin{equation}
l(x,y)=\frac{2\mu_x\mu_y + C_1}{\mu_x^2 + \mu_y^2 +C_1},
\end{equation}
\begin{equation}
c(x,y) = \frac{2\sigma_x\sigma_y + C_2}{\sigma_x^2 + \sigma_y^2 +C_2},
\end{equation}
\begin{equation}
s(x,y) = \frac{\sigma_{xy} + C_3}{\sigma_x\sigma_y +C_3},
\end{equation}
where $\mu_x$ and $\mu_y$ are the mean values, $\sigma_x$ and $\sigma_y$ are the standard deviations, and $\sigma_{xy}$ is the cross-covariance for two images $x$ and $y$.

CLIP measures the alignment between images and textual descriptions. CLIP can be defined as:
\begin{equation}
    \text{CLIPScore}(x,t) = max(cos(x,t),0)
\end{equation}
where $x$ is the visual CLIP embedding for an image, and $t$ is the textual CLIP embedding for a description. 
By evaluating the cosine similarity between the image and text embeddings, we can assess how well the reconstructed images preserve the semantic content described by the text.

LPIPS evaluates the perceptual similarity between the original and reconstructed images. It uses deep features extracted from neural networks to compare image patches, providing a perceptually motivated metric. LPIPS can be defined as 
\begin{equation}
    d_L(x,y) = \sum_l \frac{1}{W_lH_l} \sum_{h,w}\Vert w_l \odot (o_{hw}^l(x)-o_{hw}^l(y)) \Vert_2^2,
\end{equation}
where $o_{hw}^l(x)$ and $o_{hw}^l(y)$ are the activations of a neural network scaled by vector $w_l$. Then computes the normalized $\ell_2$ norm of the weighted activations.
Lower LPIPS values indicate higher perceptual similarity.

\subsection{Experimental Setups}
During our experiments, for the training, we employed AdamW ~\cite{kingma2014adam} as the optimizer, with the learning rate and weight decay set to 1e-4 and 1e-6, respectively. We utilize the SPHINX \cite{lin2023sphinx} as our LLM-based model for fine-tuning, and we initialize the two ControlNet model parameters from \cite{zhang2023adding}. Both batch sizes are set to 4 and the model is trained until it converges. The original image sizes $C, H, K$ are all set to 512, The key region image sizes $C^k, H^k, K^k$ are all set to 128, and the highly compressed image sizes $C^{ci}, H^{ci}, K^{ci}$ are all set to 32. When inference, the $\alpha$ is set to 0.5. 
In the underwater acoustic communication channel,  \(\sigma\), is set at 1.5, \(s\) at 0.5, and \(v\) at 2 m/s. Depending on the SNR, the frequency \(f\) and the distance \(d\) are adjusted to maintain the desired ratio with respect to the original signal.

\begin{table}[]
 \caption{The payload size of different transmitted data.}
 \label{table1}
\centering
\small
\begin{tabular}{c@{\hskip 3pt}|@{\hskip 3pt}c@{\hskip 3pt}|@{\hskip 3pt}c@{\hskip 3pt}|@{\hskip 3pt}c}
\hline
\textbf{\makecell{ Transmitted\\ data}} & \textbf{ Dimensionality} & \textbf{ Data Type} & \textbf{ \textbf{\makecell{Payload Size \\(Bytes $\times 10^3$)}}} \\
\hline

\makecell{ Origin image} & { {[}3,512,512{]}} & { Float 64} & { 6,291.456} \\
\hline
\makecell{ Text} & { {[}83{]}} & { Int 8} & {0.083} \\
\hline
{ \makecell{Highly \\Compressed\\ picture}} & { {[}3,32,32{]}} & { Float 64} & { 24.576} \\
\hline
\makecell{ Key Region\\ picture} & \makecell{ {[}3,64,64{]} * 0.28} & { Float 64} & { 27.525} \\
\hline
\makecell{ Proposed\\ method} & { \makecell{{[}3,32,32{]} + \\{[}3,64,64{]} * 0.28 \\+ {[}77{]}}} & { \makecell{Float 64 +\\Float 64 +\\ Int 8}} & \makecell{ 52.234 \\ \textbf{(Reduced 99.2\%)}}\\
\hline
\end{tabular}
\end{table}

Our experiments are implemented based on the PyTorch~\cite{paszke2019pytorch}. All experiments are conducted on a server running Ubuntu 20.04, equipped with a 64-core 2.60GHz Xeon(R) CPU, 256 GB of RAM, and 4 NVIDIA 4090 GPUs, each with 24 GB of memory.

\subsection{Evaluation on different approaches (\textbf{E1})}
To simulate real-world underwater scenarios, the experiments are conducted under 0, 3, 6, 9, 12, 15, 18 SNR settings.

\begin{figure*}

     \centering
     \begin{subfigure}[b]{1\textwidth}
         \centering
         \includegraphics[width=\textwidth]{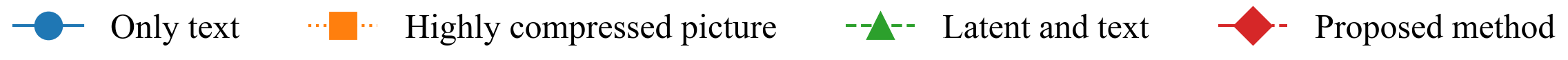}
     \end{subfigure}

\begin{subfigure}[b]{0.24\textwidth}
  \includegraphics[width=\textwidth]{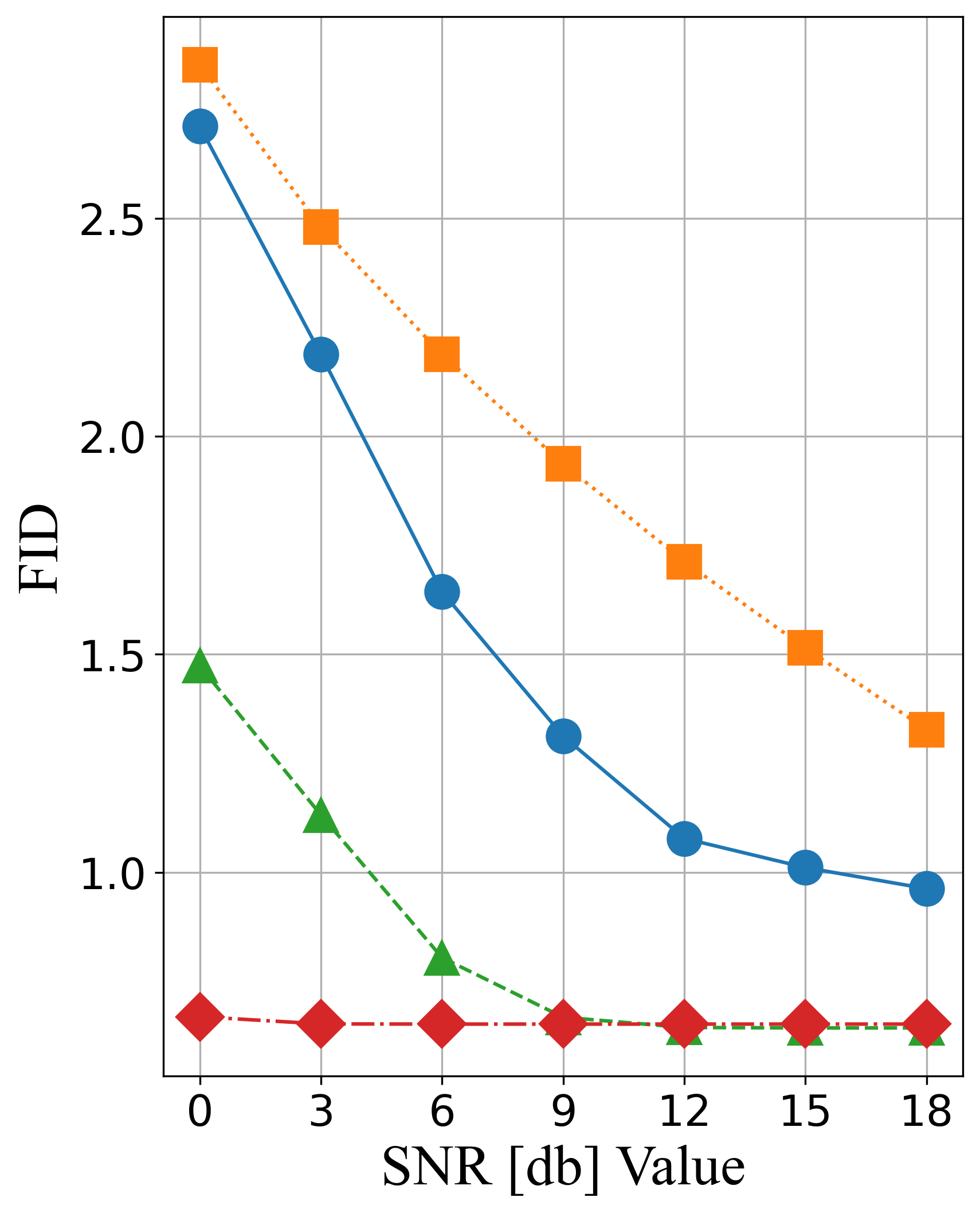}
  \caption{The testing FID}
  \label{exp1_fid}
\end{subfigure}
\begin{subfigure}[b]{0.24\textwidth}
  \centering
  \includegraphics[width=\textwidth]{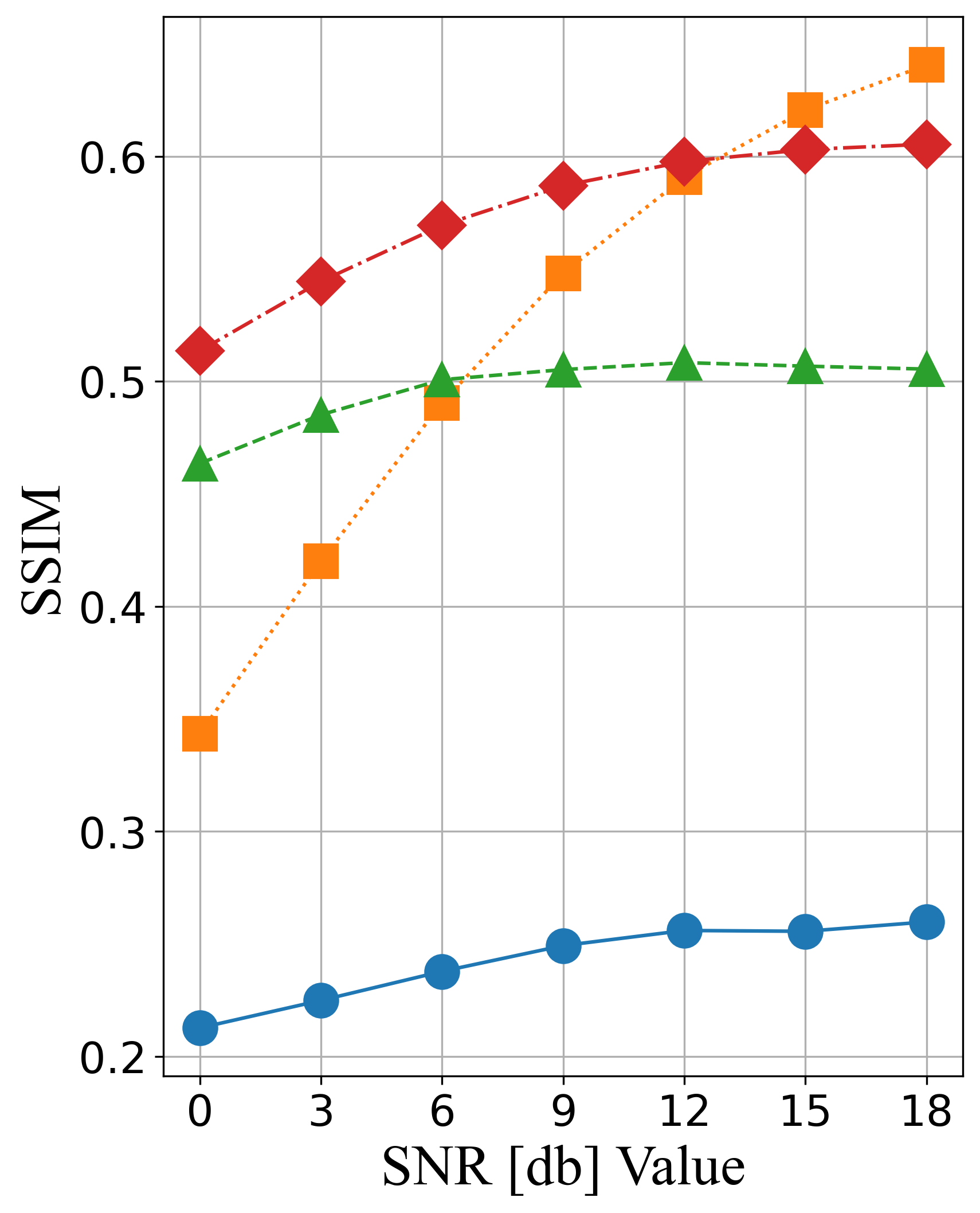}
  \caption{The testing SSIM}
  \label{exp1_ssim}
\end{subfigure}
\begin{subfigure}[b]{0.24\textwidth}
  \centering
  \includegraphics[width=\textwidth]{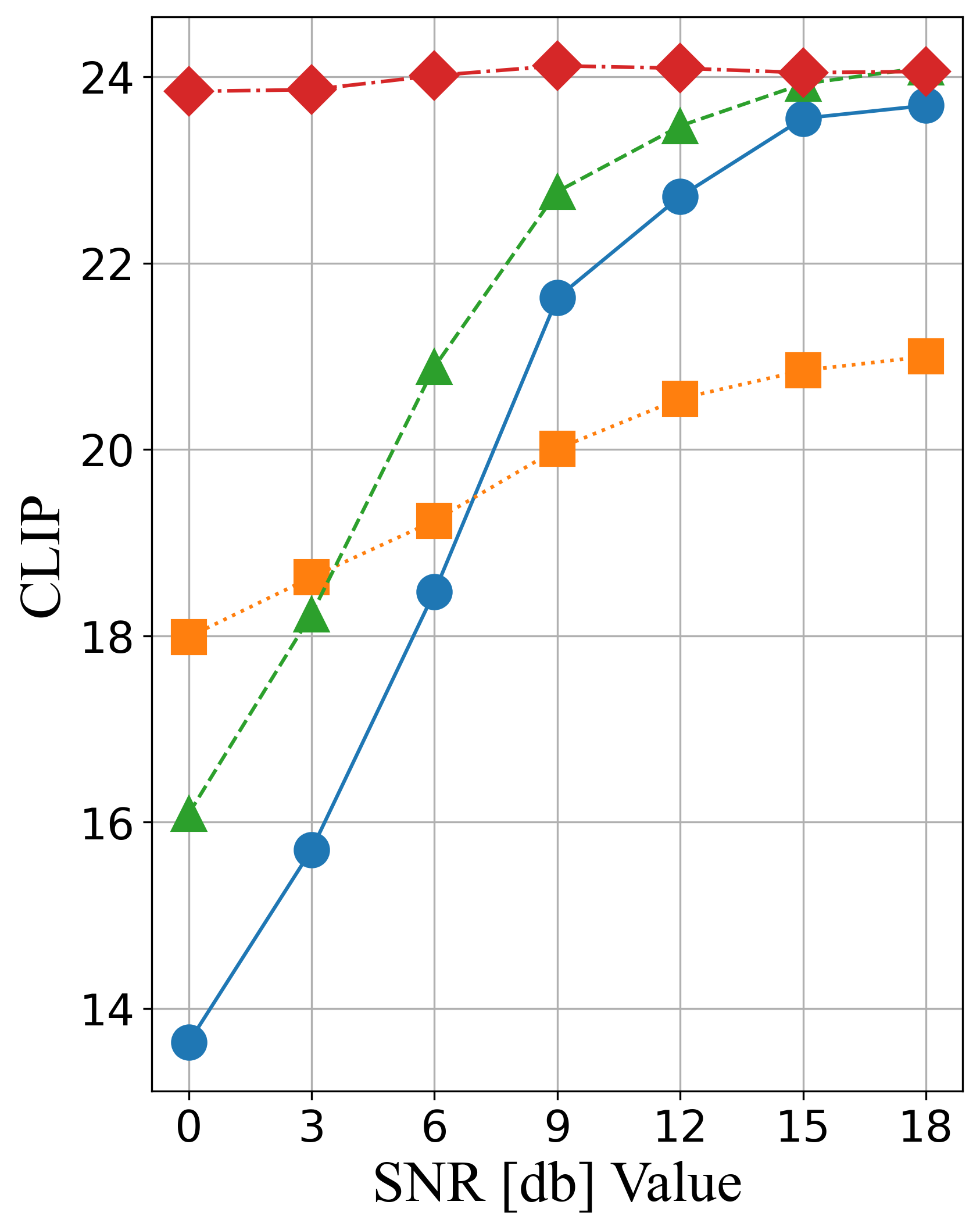}
  \caption{The testing CLIP score}
  \label{exp1_clip}
\end{subfigure}
\begin{subfigure}[b]{0.24\textwidth}
  \centering
  \includegraphics[width=\textwidth]{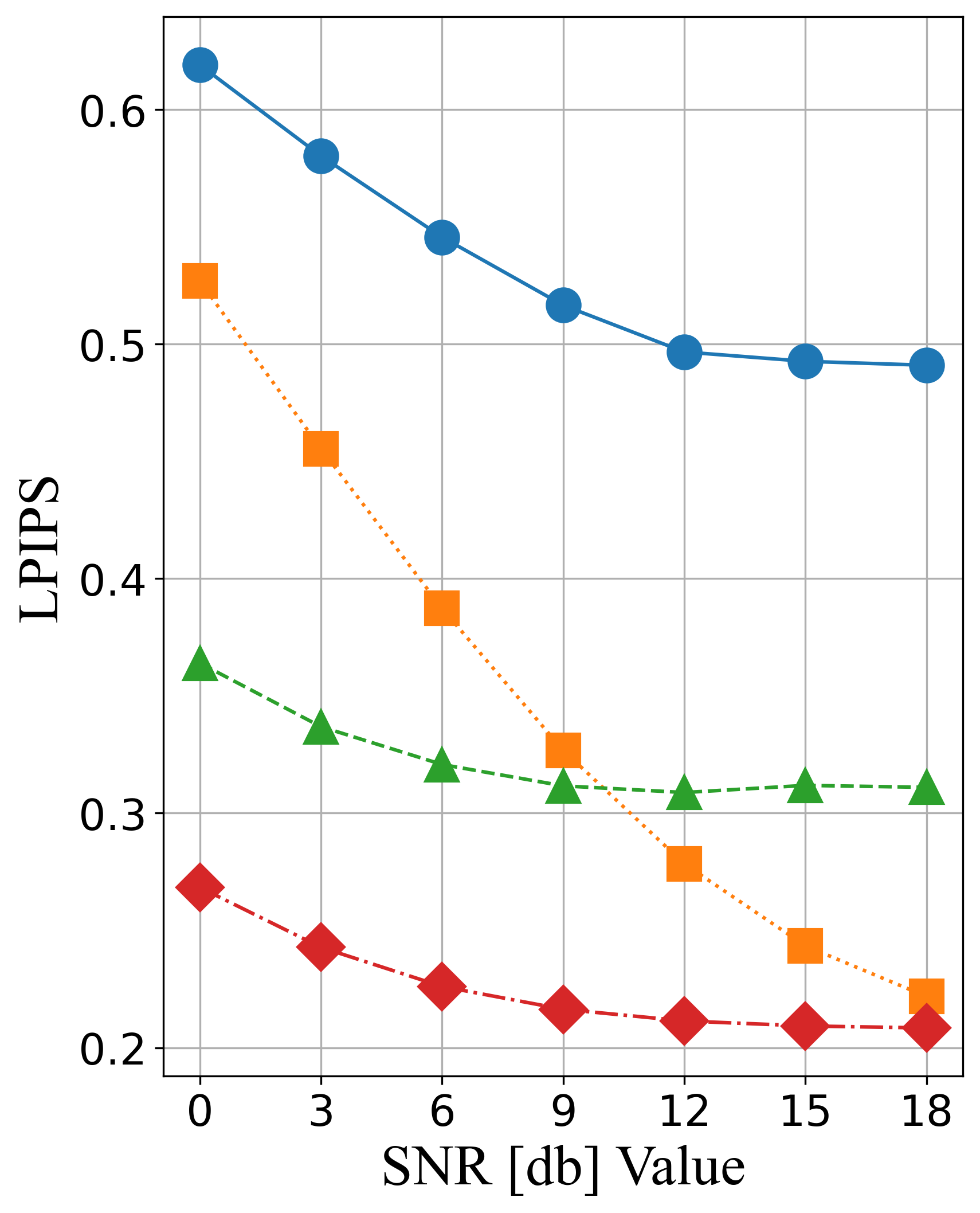}
  \caption{The testing LPIPS}
  \label{exp1_lpips}
\end{subfigure}
 \caption{The testing evaluation of different approaches.}
 \label{fig:exp1}
 \vspace{-2mm}
\end{figure*}

\begin{figure*}[ht]
    \centering
  \includegraphics[width=1\textwidth]{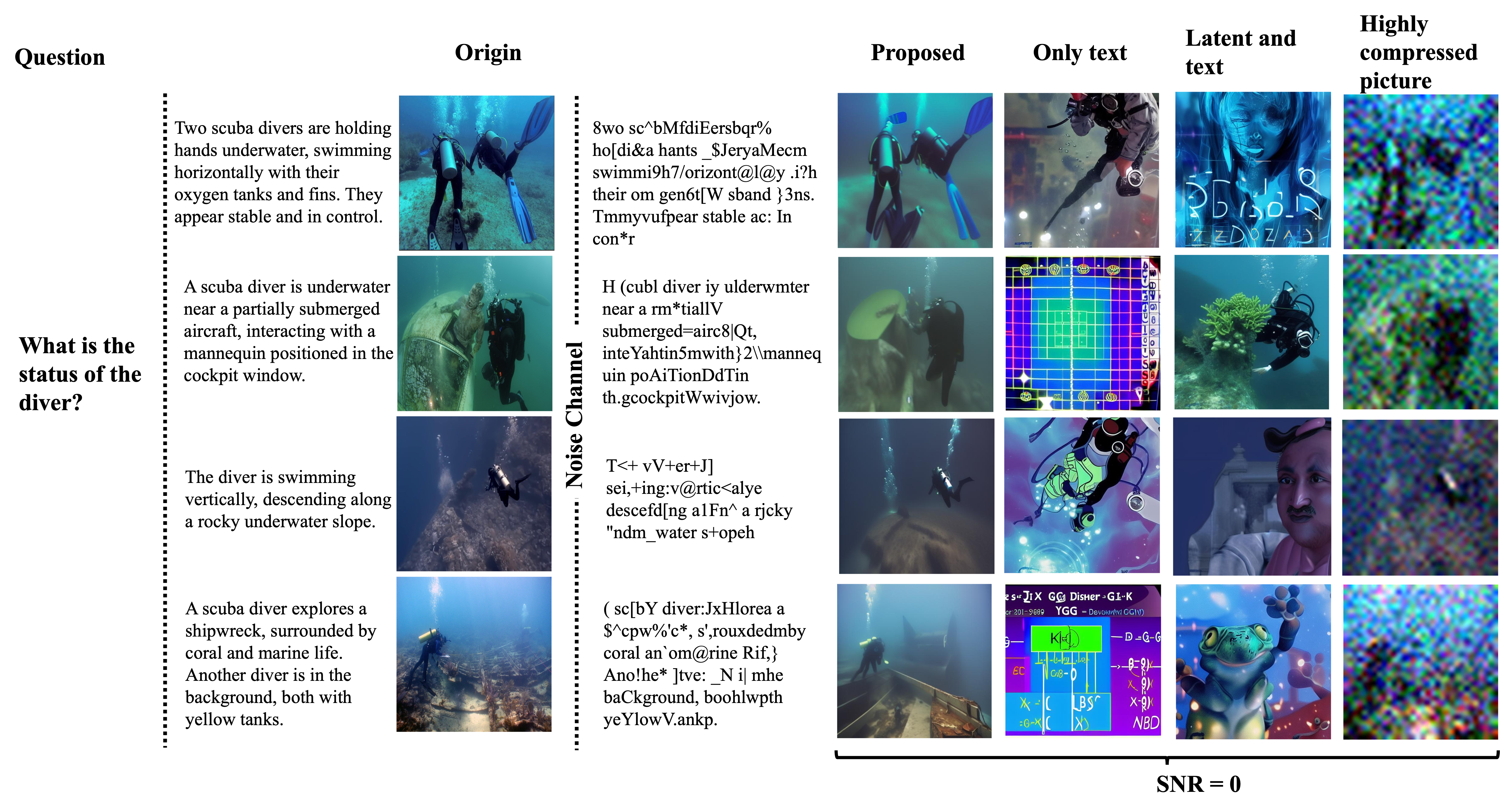}
  
  \caption{The generation picture of different approaches when SNR=0.}
  \label{generation_result}
  \vspace{-2 mm}
\end{figure*}

\subsubsection{Baselines for Comparison}
 
Our proposed method is compared with the following Three different approaches:
\begin{itemize}
 \item \textbf{Only text}: Transmit only text data and recover with text.
 \item \textbf{Highly compressed picture}: Transmit only highly compressed picture and recover with picture resampling method.
 \item \textbf{Latent and text}: a novel language-oriented SC framework that communicates both text and a compressed image embedding \cite{cicchetti2024language}. 
\end{itemize}

As shown in Table \ref{table1}, the proposed method achieves a substantial reduction in payload size, decreasing it by 99.2\% compared to the original image. On average, key regions constitute 28\% of the total image size, and the vLLM generates an average of 83 characters.
To show the performance under different noise scenarios, we evaluate the different metrics under different SNR settings, as shown in Fig. \ref{fig:exp1}. As depicted in Fig. \ref{exp1_fid}, the proposed method consistently outperforms the baseline approaches in terms of FID scores, demonstrating its ability to generate high-quality, realistic images. Lower FID scores indicate better visual quality, and our method maintains a significant margin over the others, particularly at lower SNR levels. The high visual quality is due to the effective combination of key region emphasis and holistic image reconstruction. The Highly compressed picture method performs the worst because heavy compression degrades image quality, and it cannot be recovered without diffusion methods.

Fig. \ref{exp1_ssim} illustrates that the proposed method attains the highest SSIM scores when SNR is less than 12, signifying excellent structural similarity with the original images in the highly noisy environment. The high SSIM scores are due to the method's focus on preserving both the key regions and the overall image structure. the SSIM for only transmitting the text do not significantly increase with higher SNR values because transmitting only textual information lacks the direct structural cues present in the visual data, making it inherently challenging to reconstruct the original image's structural details accurately. As a result, even as the noise decreases (higher SNR), the structural similarity cannot substantially improve because the text alone does not provide sufficient information to fully restore the image's spatial and structural attributes. The Highly compressed picture method outperforms the proposed method at higher SNR values due to its simpler approach, which benefits more from reduced noise and better preserves certain structural details. 

\begin{figure*}

     \centering
     \begin{subfigure}[b]{1\textwidth}
         \centering
         \includegraphics[width=\textwidth]{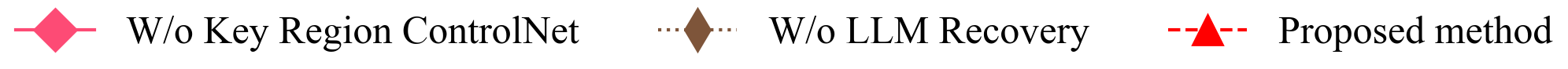}
     \end{subfigure}

\begin{subfigure}[b]{0.24\textwidth}
  \includegraphics[width=\textwidth]{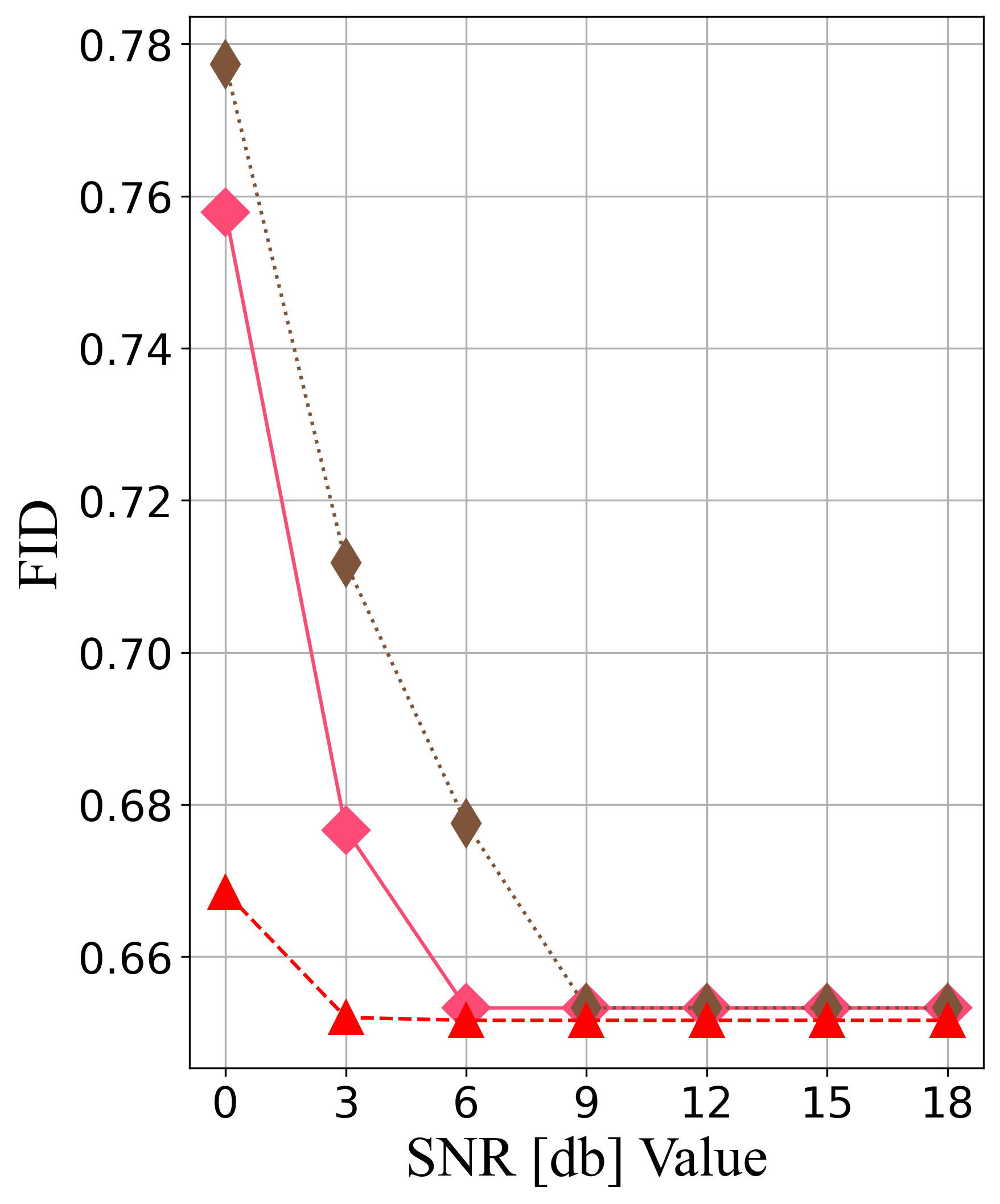}
  \caption{The testing FID}
  \label{exp2_fid}
\end{subfigure}
\begin{subfigure}[b]{0.24\textwidth}
  \centering
  \includegraphics[width=\textwidth]{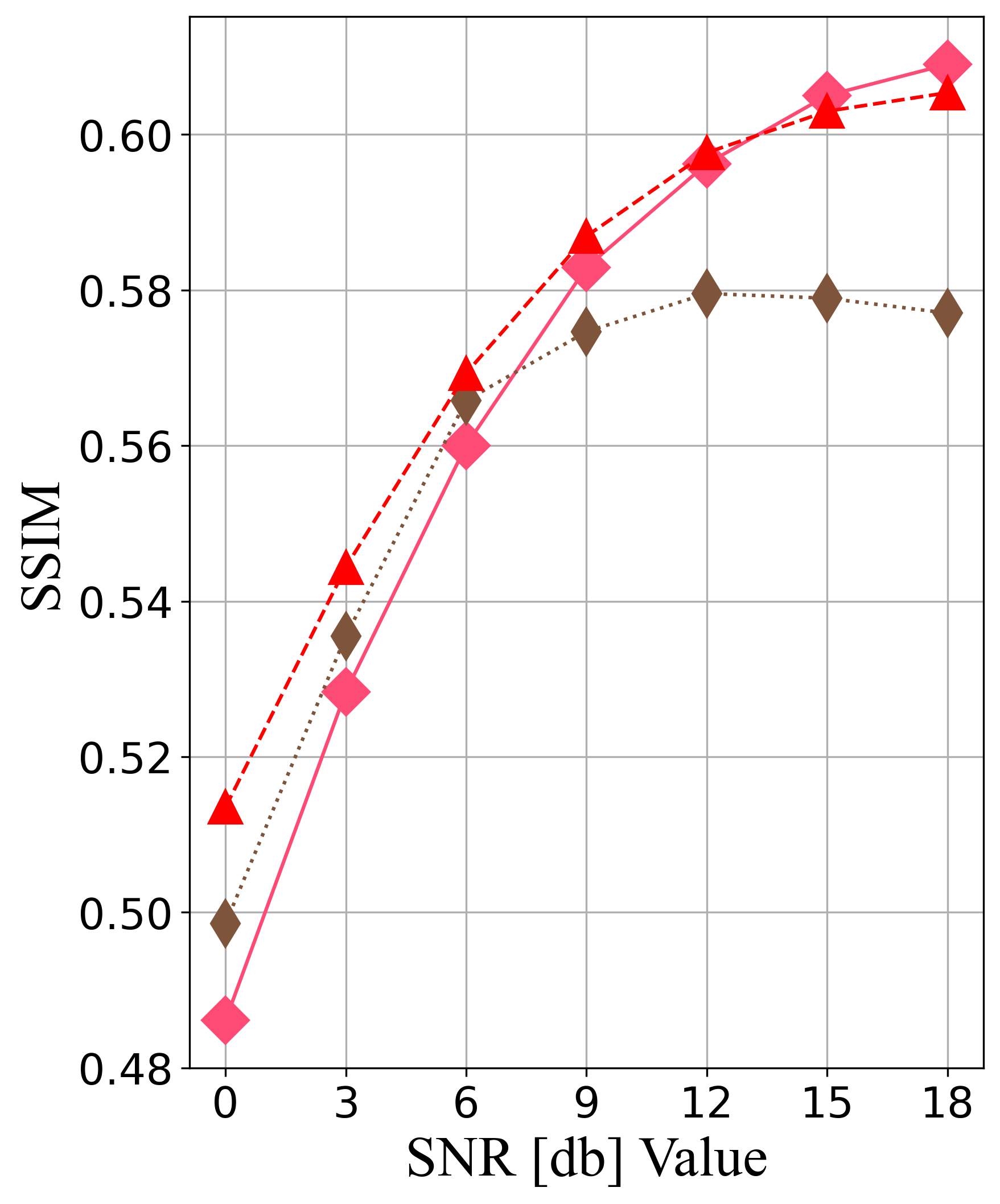}
  \caption{The testing SSIM}
  \label{exp2_ssim}
\end{subfigure}
\begin{subfigure}[b]{0.24\textwidth}
  \centering
  \includegraphics[width=\textwidth]{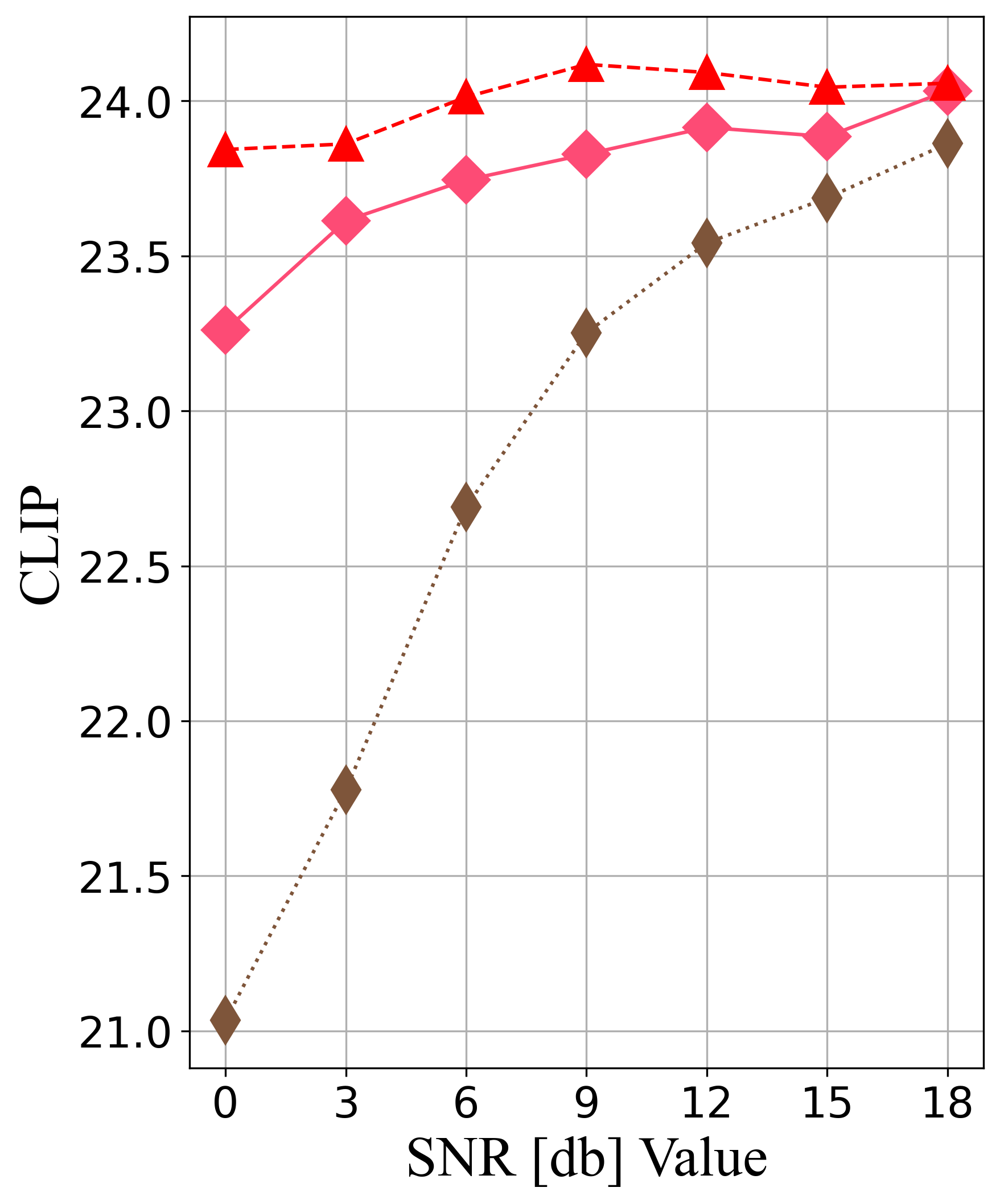}
  \caption{The testing CLIP score}
  \label{exp2_clip}
\end{subfigure}
\begin{subfigure}[b]{0.24\textwidth}
  \centering
  \includegraphics[width=\textwidth]{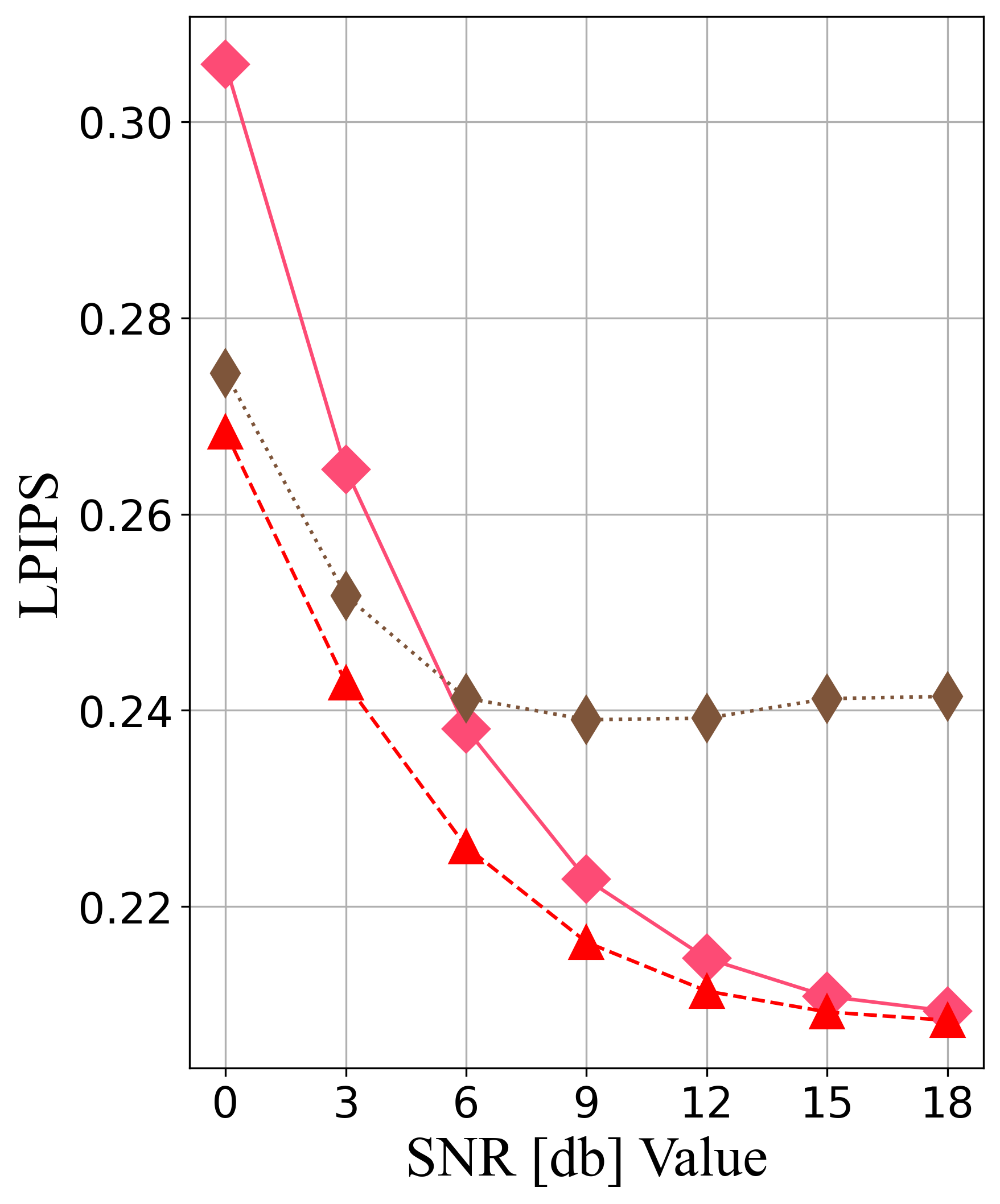}
  \caption{The testing LPIPS}
  \label{exp2_lpips}
\end{subfigure}
 \caption{The testing evaluation of ablation study.}
 \label{fig:exp1}
 \vspace{-2mm}
\end{figure*}
Fig. \ref{exp1_clip} shows that our proposed method achieves the highest CLIP scores under a more noisy environment, indicating superior semantic alignment between the reconstructed images and the textual descriptions. The high CLIP scores for the proposed method are attributed to the effective integration of textual guidance and image reconstruction, ensuring that the reconstructed images closely match the input queries. The Highly compressed picture transmission does not show a steep increase in CLIP scores like the other methods because CLIP primarily measures the alignment between text and images. Since this method does not utilize textual guidance, it cannot leverage the reduction in textual noise to improve semantic alignment as effectively as the other methods. Additionally, the Latent and text transmission surpasses the proposed method at SNR = 18 because it transmits more information, with the transmission size being 1.3 times that of the proposed method.

The LPIPS scores presented in Fig. \ref{exp1_lpips} reveal that the proposed method achieves the lowest values across all SNR settings, indicating superior perceptual similarity to the original images. The low LPIPS scores for the proposed method are a result of its ability to preserve fine details and perceptual features during the reconstruction process. The only text transmission exhibits a relatively flat LPIPS curve as SNR increases. This flatness occurs because the reconstruction process relies solely on textual information, which lacks the direct visual features needed to accurately reproduce perceptual details. As a result, even when noise is reduced (higher SNR), the lack of detailed visual cues prevents significant improvement in perceptual similarity.

The image illustrating the results of our proposed generation method is presented in \ref{generation_result}, corresponding to the scenario where $SNR=0$. Despite the heavy noise environment, the proposed method demonstrates superior image reconstruction capability. This effectiveness is due to the innovative use of LLM-based semantic communication principles, where key semantic elements are preserved while unnecessary details are effectively compressed. This approach not only enhances the image's clarity and detail in harsh underwater communication scenarios but also maintains a high level of structural and semantic integrity.

Overall, the results clearly demonstrate the effectiveness of our proposed method in achieving high-quality, semantically accurate, and perceptually similar image reconstructions under various SNR settings, significantly outperforming the baseline approaches.

\subsection{Ablation study (\textbf{E2})}
In this subsection, we systematically evaluate the impact of different components of our proposed Semantic Communication (SC) framework. Understanding the contribution of each component provides valuable insights into the robustness and adaptability of the framework. For this evaluation, we individually remove the Key Region ControlNet and the LLM recovery mechanism in the semantic decoder to assess their respective contributions. Specifically, removing the Global Vision ControlNet poses a unique challenge as the Key Region ControlNet alone is designed to recover only key regions. Thus, it would be inappropriate to measure the overall performance using standard metrics without the Global Vision ControlNet. To address this, we analyze the hyperparameters in the next section to demonstrate the impact of the Global Vision ControlNet comprehensively. 




\subsubsection{Evaluation Results}
Fig. \ref{exp2_fid} presents the FID scores for the different configurations. The absence of the Key Region ControlNet and the LLM recovery part results in significantly higher FID scores under high noise conditions, indicating that these components are crucial for maintaining image quality in noisy environments. As the SNR values increase, the FID scores for all methods converge and exhibit less variation because the reduction in noise diminishes the adverse effects of missing components.

Fig. \ref{exp2_ssim} shows that the removal of key components leads to a noticeable decrease in SSIM scores, underscoring their importance in preserving structural details in the reconstructed images. Notably, the SSIM scores for the configuration without LLM recovery do not increase as rapidly as those for the proposed method and the configuration without the Key Region ControlNet when the noise decreases. This phenomenon can be attributed to the fact that more accurate text information significantly aids in recovering structural similarity under low noise conditions.

Fig. \ref{exp2_clip} shows that the absence of the LLM recovery part leads to a significant drop in CLIP scores under high noise conditions, underscoring the crucial role of LLM recovery in restoring semantically relevant information. In contrast, removing the Key Region ControlNet results in a less pronounced, yet still noticeable, decline in CLIP scores. This indicates that while the reconstruction of key regions is less critical than LLM recovery for maintaining semantic alignment, it still provides considerable support in preserving language-related coherence.

Fig. \ref{exp2_lpips} presents the LPIPS scores for the different configurations. Similar to the SSIM scores, the removal of key components results in higher LPIPS scores under high noise conditions, whereas under low noise conditions, the absence of the LLM recovery has a more pronounced negative impact. In high noise environments, the Key Region ControlNet is essential for preserving perceptual similarity by focusing on critical image regions. However, as the noise decreases, the significance of the LLM recovery becomes more apparent. The accurate textual information provided by the LLM recovery mechanism significantly enhances perceptual quality, making its absence particularly detrimental in low noise conditions. 

\begin{figure*}

     \centering
     \begin{subfigure}[b]{0.8\textwidth}
         \centering
         \includegraphics[width=\textwidth]{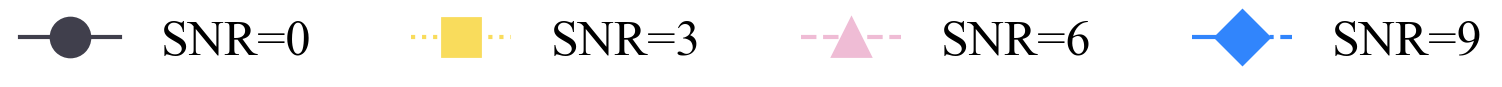}
     \end{subfigure}

\begin{subfigure}[b]{0.24\textwidth}
  \includegraphics[width=\textwidth]{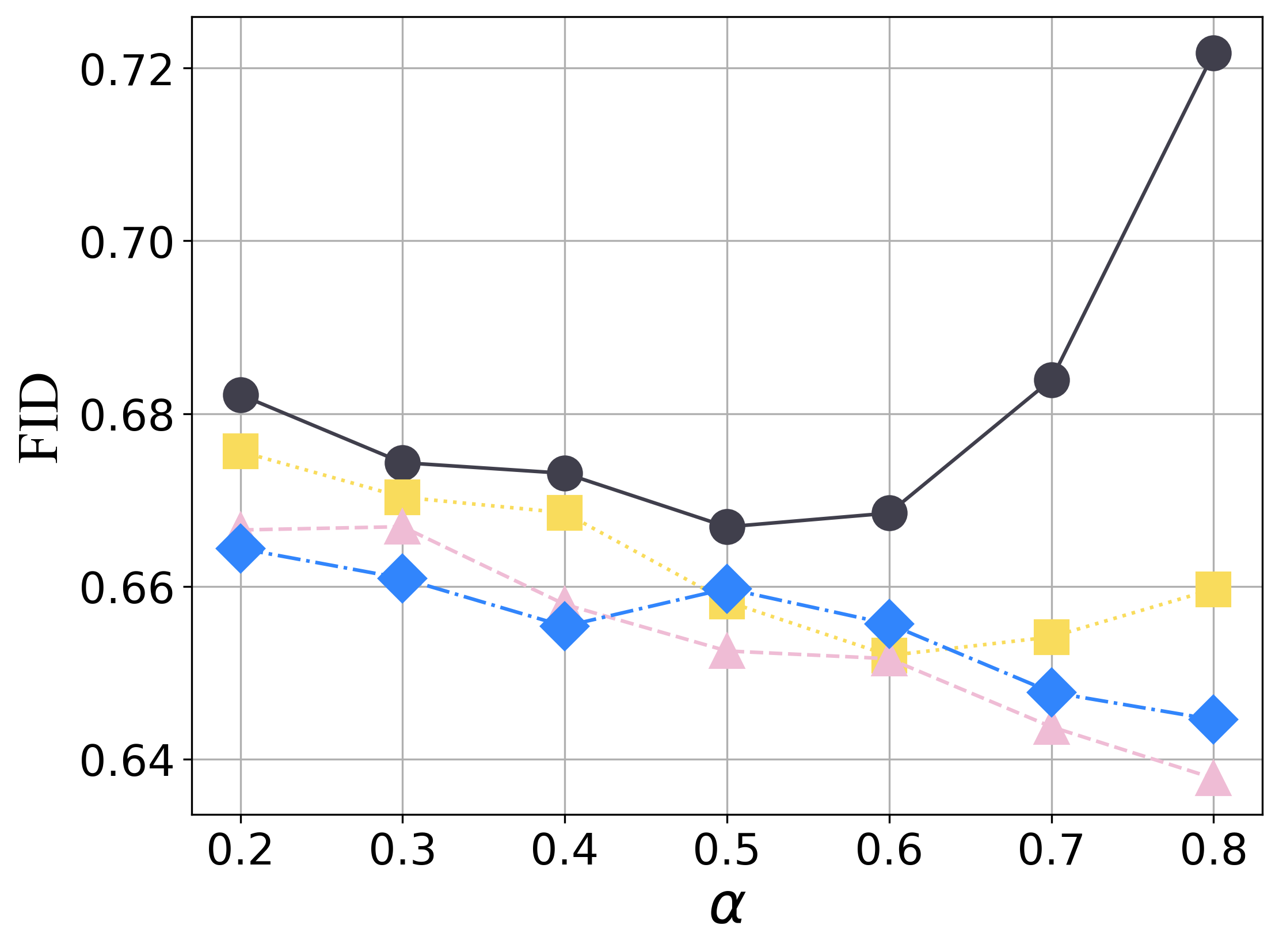}
  \caption{The testing FID}
  \label{exp3_fid}
\end{subfigure}
\begin{subfigure}[b]{0.24\textwidth}
  \centering
  \includegraphics[width=\textwidth]{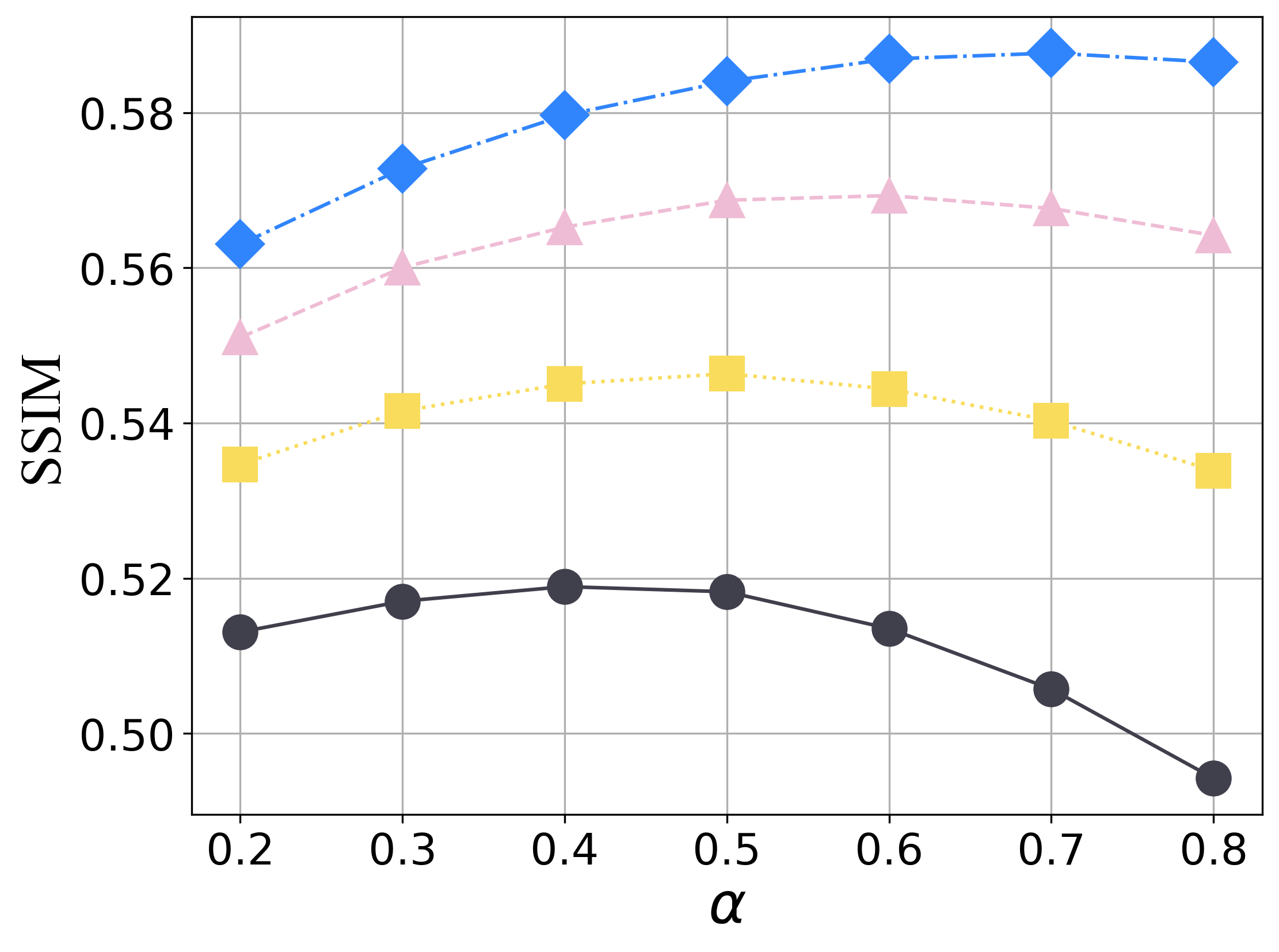}
  \caption{The testing SSIM}
  \label{exp3_ssim}
\end{subfigure}
\begin{subfigure}[b]{0.24\textwidth}
  \centering
  \includegraphics[width=\textwidth]{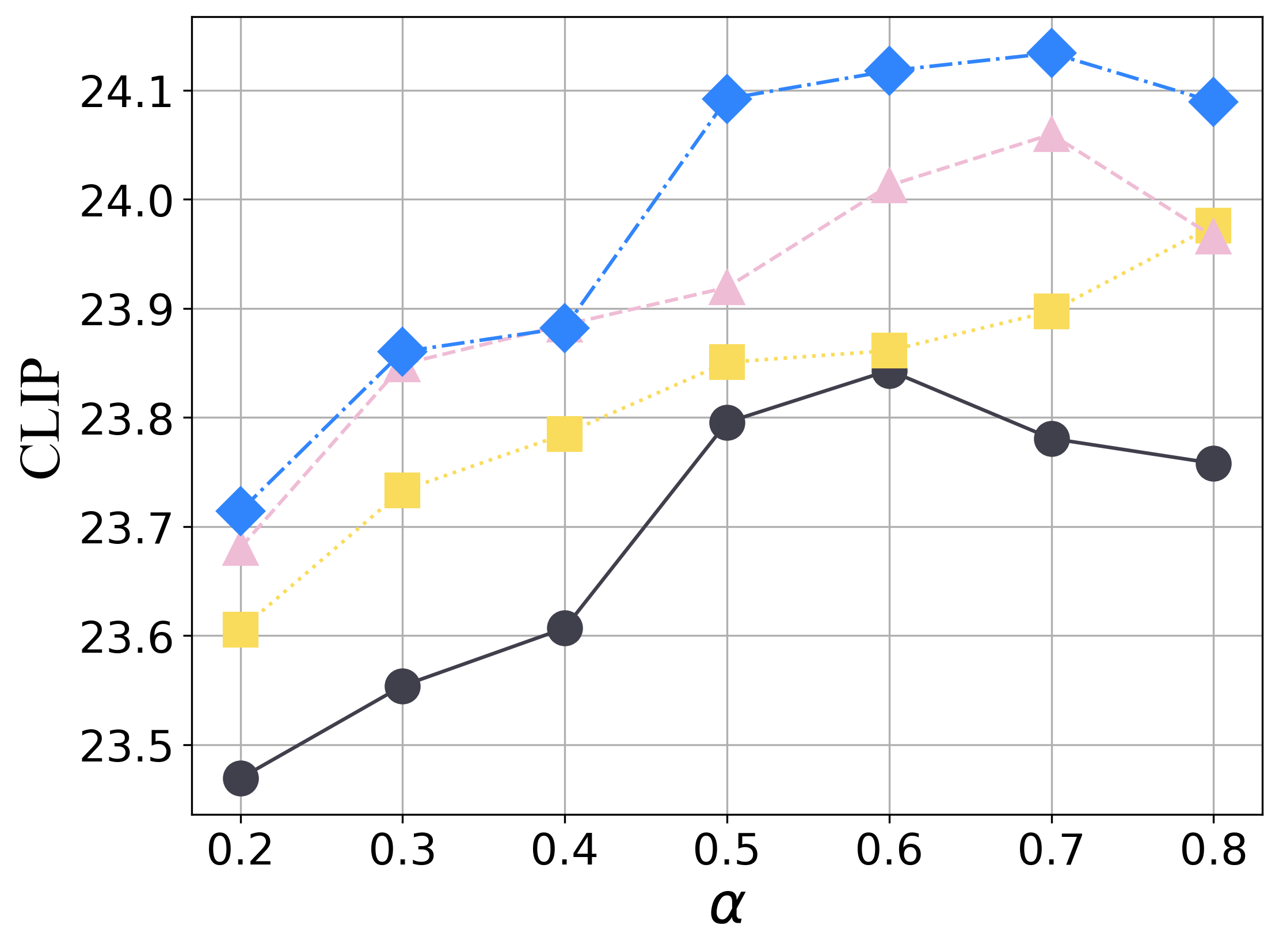}
  \caption{The testing CLIP}
  \label{exp3_clip}
\end{subfigure}
\begin{subfigure}[b]{0.24\textwidth}
  \centering
  \includegraphics[width=\textwidth]{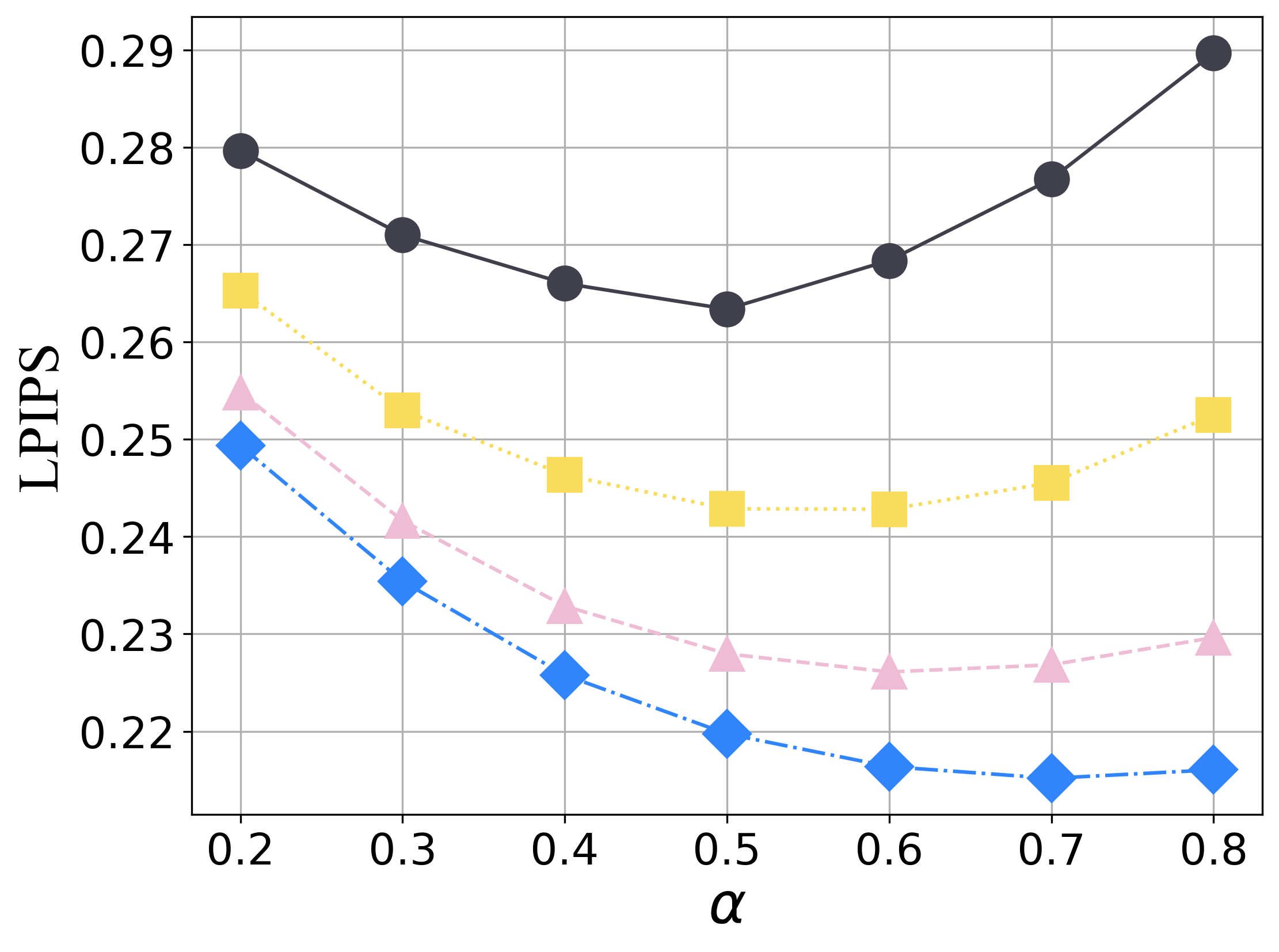}
  \caption{The testing LPIPS}
  \label{exp3_lpips}
\end{subfigure}
 \caption{The testing evaluation of different $\alpha$.}
 \label{fig:exp3}
 \vspace{-2mm}
\end{figure*}

\subsection{Evaluation on key hyperparameters (\textbf{E3})}
In this subsection, we discuss the impact of varying $\alpha$, which controls the relative influence of the Global Vision ControlNet compared to the Key Region ControlNet. We evaluate the performance of $\alpha$ values ranging from 0.2 to 0.8 under high noise conditions, specifically for SNR values from 0 to 9. The results are presented in Figure \ref{fig:exp3}.

Fig. \ref{exp3_fid} presents the FID scores for different $\alpha$ values. We observe that under extremely high noise conditions (SNR = 0 or 3), the best performance is achieved with $\alpha$ values between 0.5 and 0.6. This is because, in the presence of strong noise, the information capacity of the transmitted $\boldsymbol{f}_k$ (key regions) is greater than that of $\boldsymbol{f}_{ci}$ (contextual information), due to the larger image size of the key regions. Therefore, the impact of the Key Region ControlNet becomes more significant, leading to better results. However, as the noise decreases, the contextual information provided by $\boldsymbol{f}_{ci}$ can work more effectively, reducing the relative benefit of the key regions. Thus, the optimal $\alpha$ value shifts, and the influence of the Key Region ControlNet diminishes. Interestingly, we also observe that FID scores can become higher at lower noise levels. This phenomenon can be attributed to the nature of what FID measures: the quality and diversity of generated images compared to real images. At lower noise levels, the reconstructed images are more detailed, which can reveal more discrepancies between the generated and real images if the reconstruction is not perfect. Consequently, any small imperfections in the generated images are more noticeable, potentially leading to higher FID scores despite the lower noise. This highlights the sensitivity of FID to subtle differences in image quality and diversity, which can be more pronounced when noise is less of a factor.

Figs. \ref{exp3_ssim} and \ref{exp3_lpips} show the SSIM and LPIPS scores for different $\alpha$ values, respectively. We observe that the peak SSIM scores for all SNR settings occur around $\alpha = 0.5$ to $0.6$, indicating that both ControlNets contribute optimally to the structural similarity of the reconstructed images at these values. Similarly, the LPIPS scores exhibit a comparable trend, with the best perceptual similarity achieved near $\alpha = 0.5$ to $0.6$. These results suggest that the unique characteristics of SSIM and LPIPS, which measure structural similarity and perceptual similarity, respectively, are best satisfied when both the Global Vision ControlNet and Key Region ControlNet are balanced. The combined influence of these two ControlNets ensures that key regions and overall image context are both effectively reconstructed, leading to higher SSIM and lower LPIPS scores. This balance maximizes the retention of structural details and enhances perceptual quality, demonstrating the necessity of both ControlNets working together to achieve the best possible results.

Fig. \ref{exp3_clip} presents the CLIP scores for different $\alpha$ values. CLIP measures the similarity between images and their textual descriptions, which tend to be more generalized and descriptive. As $\alpha$ increases, the overall clarity and detail of the image improve, making it more aligned with the textual descriptions. Consequently, the CLIP scores reach their peak around $\alpha = 0.6$ to $0.7$, indicating optimal semantic alignment between the reconstructed images and the accompanying text. However, when $\alpha$ is further increased beyond this range, we observe a decline in CLIP scores. This decline suggests that an excessive focus on the Global Vision ControlNet at the expense of the Key Region ControlNet can lead to a loss of specific details that are crucial for maintaining the nuanced alignment between the images and their textual descriptions. In other words, while greater $\alpha$ values initially enhance overall image quality, overly high values may diminish the semantic coherence of key regions, thereby reducing the effectiveness of the CLIP score. This balance underscores the importance of a nuanced approach to optimizing $\alpha$ to achieve the best semantic alignment.

In summary, the evaluation of different $\alpha$ values demonstrates that a balanced contribution from the Global Vision ControlNet and the Key Region ControlNet is essential for optimizing semantic alignment, image quality, perceptual similarity, and structural fidelity under high noise conditions.

\section{Conclusion}\label{section5}
In this paper, we have introduced a novel Semantic Communication (SC) framework based on Large Language Models (LLMs) to address the challenges of underwater communication. Our framework leverages visual LLMs for semantic compression and prioritization of underwater image data, ensuring the preservation of key semantic elements while significantly reducing the overall data size. By integrating two newly designed ControlNet networks with a diffusion model, our method achieves superior image reconstruction quality, maintaining high perceptual and structural fidelity even under noisy conditions. Experiments show that our method not only enhances semantic alignment and perceptual quality but also ensures robust and efficient data transmission, reducing the data size to 3.3\% of the original. These findings highlight the potential of our approach to significantly improve the performance of underwater communication systems.

\bibliographystyle{IEEEtran}
\bibliography{ref.bib}

\end{document}